\pdfoutput=1

\documentclass[journal,twoside,nocompress,letterpaper]{IEEEtran}

\usepackage{cite}
\usepackage{graphicx}
\usepackage{multirow}
\usepackage{textcomp}
\usepackage{amsfonts}
\usepackage{amsmath}
\usepackage{amssymb}
\usepackage{amsthm} 
\usepackage{enumitem}
\usepackage{algorithm,algorithmic}
\usepackage[table]{xcolor}
\usepackage{array}
\usepackage{placeins}
\usepackage{moresize}
\usepackage[export]{adjustbox}

\ifCLASSOPTIONcompsoc
  \usepackage[caption=false,font=footnotesize,labelfont=sf,textfont=sf]{subfig}
\else
  \usepackage[caption=false,font=footnotesize]{subfig}
\fi
\usepackage{adjustbox}
\usepackage{tcolorbox}
\usepackage{diagbox}

\ifCLASSOPTIONcaptionsoff
  \usepackage[nomarkers]{endfloat}
 \let\MYoriglatexcaption\caption
 \renewcommand{\caption}[2][\relax]{\MYoriglatexcaption[#2]{#2}}
\fi

\usepackage{tabularx}
\usepackage{booktabs}                   % Nicer tables
\usepackage{multicol}                   % Content of a table over several columns
\usepackage{multirow}                   % Content of a table over several rows
\usepackage{makecell}
\usepackage{rotating}                   % Alows to rotate text and objects
\usepackage{enumitem}
\usepackage{url}

\ifCLASSINFOpdf
  \usepackage[pdftex]{thumbpdf}
\else
  \usepackage[dvips]{thumbpdf}
\fi

\newcommand\MYhyperrefoptions{bookmarks=true,bookmarksnumbered=true,
pdfpagemode={UseOutlines},plainpages=false,pdfpagelabels=true,
colorlinks=true,linkcolor={black},citecolor={green},urlcolor={blue},
pdftitle={MS-MDA},
pdfsubject={EEG-based Emotion Recognition},
pdfauthor={Hao Chen},
pdfkeywords={brain-computer interface, deep learning, EEG, emotion recognition, affective computing, domain adaptation, transfer learning}}
\ifCLASSINFOpdf
\usepackage[\MYhyperrefoptions,pdftex]{hyperref}
\else
\usepackage[\MYhyperrefoptions,breaklinks=true,dvips]{hyperref}
\usepackage{breakurl}
\fi
\usepackage[nolist]{acronym}
\usepackage{balance}

\begin{document}
\bstctlcite{IEEEexample:BSTcontrol}
{
    \title{MS-MDA: Multisource Marginal Distribution Adaptation for Cross-subject and Cross-session EEG Emotion Recognition}
    \author{Hao Chen, Ming Jin, Zhunan Li, Cunhang Fan, Jinpeng Li*, \IEEEmembership{Member, IEEE}, Huiguang He, \IEEEmembership{Senior Member, IEEE}

\thanks{This paragraph of the first footnote will contain the date on 
which you submitted your paper for review. This work was supported by the National Natural Science Foundation of China (62020106015) ,the Strategic Priority Research Program of CAS (XDB32040000), the Zhejiang Provincial Natural Science Foundation of China (LQ20F030013), and the Ningbo Public Service Technology Foundation, China (202002N3181).}
% \thanks{The next few paragraphs should contain 
% the authors' current affiliations, including current address and e-mail. For 
% example, F. A. Author is with the National Institute of Standards and 
% Technology, Boulder, CO 80305 USA (e-mail: author@boulder.nist.gov). }
\thanks{Hao Chen, Ming Jin, Zhunan Li, Jinpeng Li are with HwaMei Hospital, University of Chinese Academy of Sciences, Ningbo, Zhejiang Province, 315100, China, and are also with Ningbo Institute of Life and Health Industry, University of Chinese Academy of Sciences, Ningbo, Zhejiang Province, 315100, China. \textit{Corresponding author: Jinpeng Li (lijinpeng@ucas.ac.cn).}}
\thanks{Cunhang Fan is with Anhui Province Key Laboratory of Multimodal Cognitive Computation, School of Computer Science and Technology, Anhui University, No. 111 Jiulong Road, Shushan District, Hefei, China. \textit{email: fchbuct@126.com}}
\thanks{Huiguang He is with Institute of Automation, Chinese Academy of Sciences, No. 95 Zhongguancun East Road, Haidian District, Beijing, China. \textit{email: huiguang.he@ia.ac.cn}}
}}

\IEEEtitleabstractindextext{
    \begin{abstract}
    As an essential element for the diagnosis and rehabilitation of psychiatric disorders, the electroencephalogram (EEG) based emotion recognition has achieved significant progress due to its high precision and reliability. However, one obstacle to practicality lies in the variability between subjects and sessions. Although several studies have adopted domain adaptation (DA) approaches to tackle this problem, most of them treat multiple EEG data from different subjects and sessions together as a single source domain for transfer, which either fails to satisfy the assumption of domain adaptation that the source has a certain marginal distribution, or increases the difficulty of adaptation. We therefore propose the multi-source marginal distribution adaptation (MS-MDA) for EEG emotion recognition, which takes both domain-invariant and domain-specific features into consideration. First, we assume that different EEG data share the same low-level features, then we construct independent branches for multiple EEG data source domains to adopt one-to-one domain adaptation and extract domain-specific features. Finally, the inference is made by multiple branches. We evaluate our method on SEED and SEED-IV for recognizing three and four emotions, respectively. Experimental results show that the MS-MDA outperforms the comparison methods and state-of-the-art models in cross-session and cross-subject transfer scenarios in our settings. Codes at https://github.com/VoiceBeer/MS-MDA.
\end{abstract}


}

\maketitle

\IEEEdisplaynontitleabstractindextext
\IEEEpeerreviewmaketitle

\ifCLASSOPTIONcompsoc
\IEEEraisesectionheading{\section{Introduction}\label{sec:introduction}}
\else

\section{Introduction}
\label{sec:introduction}
Emotion as physiological information, unlike widely studied logical intelligence, is central to the quality and range of daily human communications \cite{dolan2002emotion, tyng2017influences}. In the human-computer interaction (HCI), emotion is crucial in influencing situation assessment and belief information, from cue identification to situation classification, with decision selection for building a friendly user interface \cite{jeon2017emotions}. For example, affective brain-computer interfaces (aBCIs), acting as a bridge between the emotions extracted from the brain and the computer, which has shown potential for rehabilitation and communication \cite{birbaumer2006breaking, lee2019towards, frisoli2012new}. Besides, many studies have shown a strong correlation between emotions and mental illness. Barrett~\emph{et al.} \cite{barrett2001knowing} studies the relation between emotion differentiation and emotion regulation. Joormann~\emph{et al.} \cite{joormann2010emotion} finds that depression is strongly associated with the use of emotion regulation strategies. Bucks~\emph{et al.} \cite{bucks2004emotion} investigates the identification of non-verbal communicative signals of emotion in people that are suffering from Alzheimer’s disease. To quantify emotion, most researchers have focused on using conventional methods such as classifying emotions with facial expression or language \cite{ekman1993facial}. In recent years, with the advantage of reliability, easy accessibility, and high precision, non-invasive BCIs such as electroencephalogram (EEG) are widely used for brain signal acquisition, and analysis of psychological disorders \cite{ay2019automated, sanei2013eeg, acharya2015computer, liu2015boosting}. With EEG signals, many works also investigate the rehabilitation methods for psychological disorders, such as \cite{jiang2021enhancing} of using spatial information of EEG signals to classify depressions, and Zhang~\emph{et al.} \cite{zhang2020brain} proposes a brain functional network framework for major depressive disorder by using the EEG signals. Besides, Hosseinifard~\emph{et al.} \cite{hosseinifard2013classifying} investigates the nonlinear features from EEG signals for classifying depression patients and normal subjects. The flow of an EEG-based affective BCI (aBCI) for emotion recognition is introduced in Section.~\ref{sec:diagram}

Due to the non-stationary between individual sessions and subjects of EEG signals \cite{sanei2013eeg}, it is still challenging to get a model that is shareable to different subjects and sessions in EEG-based emotion recognition scenarios, which elicits two scenarios: cross-subject and cross-session (\emph{i.e.,} data collected from the same subject at the same session can be very biased, detailed description is given in Section \ref{scenarios}). Besides, the analysis and classification of the collected signals are time-consuming and labor-intensive, so it is important to make use of the existing labeled data to analyze new signals in the EEG-based BCIs. With this purpose, domain adaptation is widely used in research works. As a sub-field of machine learning, DA improves the learning in the unlabeled target domain through the transfer of knowledge from the source domains, which can significantly reduce the number of labeled samples \cite{pan2009survey}. In practice, we often face the situation that contains multiple source domain data (\emph{i.e.,} data from different subjects or sessions). Due to the shift between domains, adopting DA for EEG data especially when facing multiple sources is difficult. In recent years, the researchers tend to merge all source domains into one single source and then use DA to align the distribution (Source-combine DA in Fig. \ref{intro}). This simple approach may improve the performance because it expands the training data for the model, but it ignores the non-stationary of each EEG source domain itself and disrupts it (\emph{i.e.,} EEG data of different people obey different marginal distributions), besides, directly merging into one new source domain cannot determine whether its new marginal distribution still obeys EEG-data distribution, thus brings a larger bias.
 
\begin{figure}[!t]
\centerline{\includegraphics[width=\columnwidth]{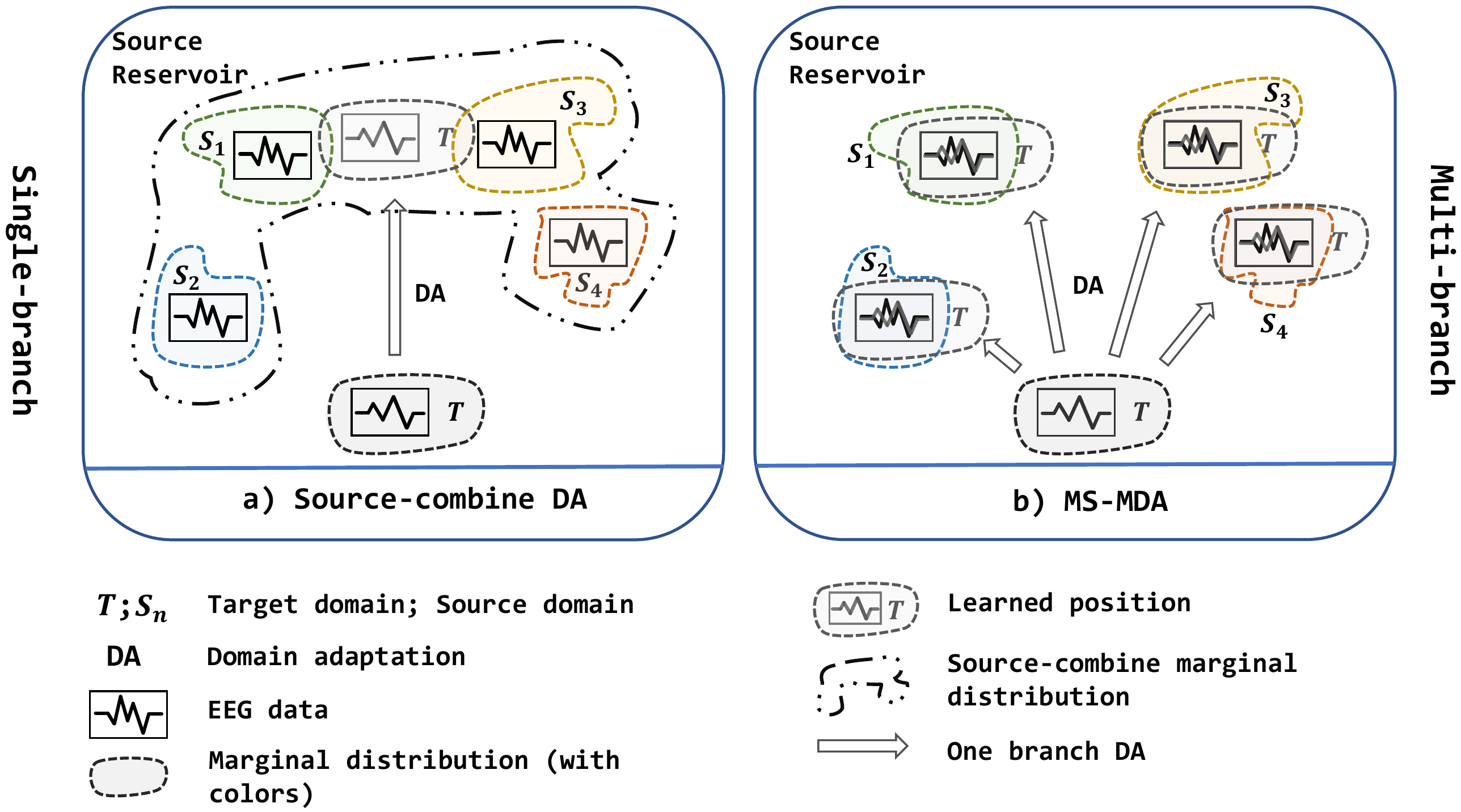}}
\caption{Two strategies of multi-source domain adaptation. a) is a single-branch strategy while b) is a multi-branch strategy. In a), all source domains are combined into one new big source and then been used to align distribution with the target domain, while in b), multiple sources are being aligned at the same time, and are divided into multiple branches to adopt DA with the target domain. In short, a) is one source, one branch with one-to-one DA; b) is multiple sources, multiple branches with one-to-one DA. The figure is best viewed in color.}
\label{intro}
\end{figure}

To solve the multi-source domain adaptation problems in EEG-based emotion recognition, we propose a multi-source marginal distribution adaptation for cross-subject and cross-session EEG emotion recognition (MS-MDA, as illustrated in Fig. \ref{intro}). First, we assume \textit{all the EEG data share low-level features, especially those taken from the same device, the same subject and the same session}. Based on this, we construct a simple common feature extractor to extract domain-invariant features. Then for multiple sources, since \textit{each of them has some specific features}, we pair every single source domain with the target domain to form a branch for one-to-one DA, and align the distribution and extract domain-specific features. After that, a classifier is trained for each branch, and \textit{the final inference is made by these multiple classifiers from multiple branches}. The details of MS-MDA are given in Section \ref{method}. 

In summary, we make three following contributions:

\begin{enumerate}
    \item We propose MS-MDA for EEG-based emotion recognition in a new multi-source adaptation way to avoid disrupting the marginal distributions of EEG data.
    \item Extensive experiments demonstrate that our method outperforms the comparison methods on SEED and SEED-IV, and additional experiments also illustrate that our method generalizes well.
    \item During the experiments, we also notice the importance of normalizing the EEG data, thus we design and evaluate few normalization approaches for EEG data in the domain adaptation scenarios and draw corresponding conclusions. To our knowledge, we are the first to investigate the normalization methods for EEG data, which we believe can be taken as a guide for other future works, and be applied to all data in EEG-based datasets and EEG-related domains.
\end{enumerate}

In the remainder of this paper, we first review related works on domain adaptation in the field of EEG-based emotion recognition in Section \ref{related_work}. Section \ref{materials} introduces the materials, including the diagram of EEG-based affective BCI with transfer scenarios, datasets and pre-processing methods. The details of MS-MDA are given in Section \ref{method}, whereas Section \ref{experiments} demonstrates the settings, results, and additional experiments. Section \ref{conclusion} discusses the results of the experiment and our findings, as well as problems and solutions. Finally, Section \ref{conclusion} concludes the work and outlines the future extension.

\section{Related Work}\label{related_work}
In recent years, the research of affective computing has become one of the trends of machine learning, neural systems, and rehabilitation study. Among those works, emotions are usually characterized into two types of emotion model: discrete categories (basic emotional states, \emph{e.g.,} happy, sad, neutral \cite{zheng2015investigating}) or continuous values (\emph{e.g.,} in 3D space of arousal, valence and dominance \cite{koelstra2011deap}). With domain adaptation techniques, many works have achieved significant performance in the field of affective computing.

Zheng~\emph{et al.} \cite{zheng2016personalizing} first applies Transfer Component Analysis \cite{pan2010domain} and Kernel Principle Analysis based methods on SEED dataset to personalize EEG-based affective models and demonstrates the feasibility of adopting DA in EEG-based aBCIs. Chai~\emph{et al.} proposes adaptive subspace feature matching \cite{chai2017fast} to decrease the marginal distribution discrepancy between two domains, which requires no labeled samples in the target domain. To solve cross-day binary classification, Lin~\emph{et al.} \cite{lin2017improving} extends robust principal component analysis (rPCA) \cite{candes2011robust} to their filtering strategy which can capture EEG oscillations of relatively consistent emotional responses. Li~\emph{et al.}, different from the above, considering the multi-source scenario, and proposes a Multi-source Style Transfer Mapping (MS-STM) \cite{li2019multisource} framework for cross-subject transfer. They first take a few labeled training data to learn multiple STMs, which are then being used to map the target domain distribution to the space of the sources. Their method is similar to our MS-MDA, but they do not take the domain-invariant features into consideration, thus losing the low-level information.

In recent years, with the development of deep learning techniques and its usability, many works of EEG-based decoding with neural networks have been proposed. Jin~\emph{et al.} \cite{jin2017eeg}, and Li~\emph{et al.} \cite{li2018cross} adopts deep adaptation network (DAN) \cite{long2015learning} to EEG-based emotion recognition, which takes maximum mean discrepancy (MMD) \cite{borgwardt2006integrating} as a measure of the distance between the source and the target domain, and training to reduce it on multiple layers. Extending the original method, Chai~\emph{et al.} proposes subspace alignment auto-encoder (SAAE) \cite{chai2016unsupervised} which first projects both source and target domains into a domain-invariant subspace using an auto-encoder, and then kernel PCA, graph regularization and MMD are used to align the feature distribution. To adapt the joint distribution, Li~\emph{et al.} \cite{li2019domain} propose a domain adaptation method for EEG-based emotion recognition by simultaneously adapting marginal distributions and conditional distributions, they also present a fast online instance transfer (FOIT) for improved EEG emotion recognition \cite{li2020foit}. Zheng~\emph{et al.} extends SEED dataset to SEED-IV dataset and presents EmotionMeter \cite{zheng2018emotionmeter}, a multi-modal emotion recognition framework that combines two modalities of eye movements and EEG waves. With the concept of attention-based convolutional neural network (CNN) \cite{yin2016abcnn}, Fahimi~\emph{et al.} \cite{fahimi2019inter} develops an end-to-end deep CNN for cross-subject transfer and fine-tunes it by using some calibration data from the target domain. To tackle the requirement of amassing extensive EEG data, Zhao~\emph{et al.} \cite{zhao2021plug} proposes a plug-and-play domain adaptation method for shortening the calibration time within a minute while maintaining the accuracy. Wang~\emph{et al.} \cite{wang2021prototype} present a domain adaptation SPD matrix network (daSPDnet) to help cut the demand of calibration data for BCIs.

These aBCI works have gained significant improvement in their respective directions, transfer scenarios, and on multiple benchmark databases. However, many of them focus on combing multiple sources into one and adopt one-to-one DA, which ignores the differences of the marginal distribution of different EEG domains (source-combine DA in Fig. \ref{intro}). This operation may compromise the effectiveness of downstream tasks, and although it somehow extends the training data, the trained models do not generalize well enough. Therefore, inspired by \cite{zhu2019aligning}, a novel multi-source transfer framework, we propose {\bf MS-MDA} (multi-source marginal distribution alignment for EEG-based emotion recognition), which transfers multiple source domains to the target domain separately, thus avoiding the destruction of the marginal distribution of the multiple EEG source domains; and also takes the domain-invariant features into consideration. Due to the sensitivity of the EEG data and intuition, we do not adopt complex networks, but just a combination of few multi-layer perceptrons (MLPs) \cite{gardner1998artificial}, and thus makes our method computationally efficient, and easy to expand.

\section{Materials}\label{materials}

\subsection{Diagram}\label{sec:diagram}
\begin{figure}[!t]
\centerline{\includegraphics[width=\columnwidth]{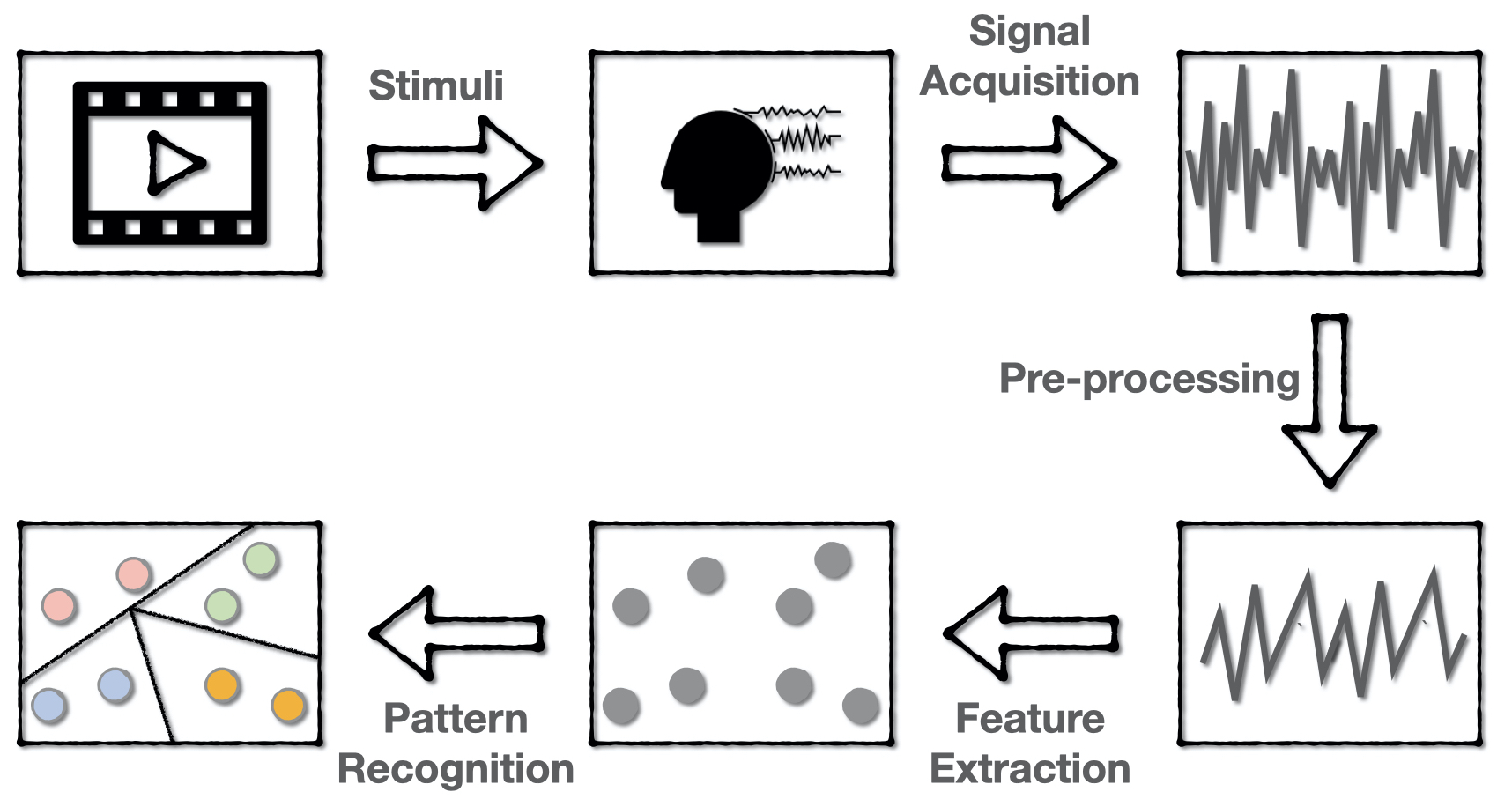}}
\caption{The {\bf flowchart} of EEG-based BCI for emotion recognition. The emotions are first evoked and encoded into EEG data, then the EEG data are pre-processed and extracted to various forms of features for subsequent pattern recognition.}
\label{eeg_diagram}
\end{figure}

The flow of one EEG-based aBCI for emotion recognition is shown in Fig. \ref{eeg_diagram}, which involves five steps:

\begin{itemize}
    \item {\bf Stimulating emotions.} The subjects are first stimulated with stimuli that correspond to a target emotion. The most commonly used stimuli are movie clips with sound, which can better stimulate the desired emotion because they mix sound with images and actions. After each clip, self-assessment is also applied for the subject to ensure the consistency of the evoked emotion and the target emotion.
    \item {\bf EEG signal acquisition and recording.} The EEG data are collected using the dry electrodes on the BCI, and then be labeled with the target emotion.
    \item {\bf Signal pre-processing.} Since the EEG data is a mixture of various kinds of information containing much noise, it is required to pre-process the EEG signal to get cleaner data for subsequent recognition. This step often includes down-sampling, band-pass filtering, temporal filtering, and spatial filtering to improve the signal-to-noise ratio (SNR).
    \item {\bf Feature extraction.} In this step, features of the pre-processed signals are extracted in various ways. Most of the current research works are to extract features in the time or frequency domain.
    \item {\bf Pattern recognition.} The use of machine learning techniques to classify or regress data according to specific application scenarios.
\end{itemize}

\subsection{Scenarios}\label{scenarios}
Considering the sensitivity of the EEG, domain adaptation in emotion recognition can be divided into several cases: 1) \textbf{Cross-subject transfer.} In one session, new EEG data from a new subject is taken as the target domain, and the rest of existing EEG data from other subjects are taken as the source domains for DA. 2) \textbf{Cross-session transfer.} For one subject, data collected in the previous sessions can be used as the source domain for DA, and data collected in the new session are taken as the target domain.

In our work, since the datasets we evaluate on contains 3 session and 15 subjects (refer to Section \ref{datasets} for details), we take the first 2 session data from one subject as the source domains for cross-session transfer, and take the first 14 subjects data from one session as the source domains for cross-subject transfer. The results of cross-session scenarios are averaged over 15 subjects, and the results of cross-subject are averaged over 3 sessions. Standard deviations are also calculated.

\subsection{Datasets}\label{datasets}
The database we evaluate on are: {\bf SEED} \cite{zheng2015investigating} \cite{duan2013differential} and {\bf SEED-IV} \cite{zheng2018emotionmeter}, both are established by the BCMI laboratory led by Prof. Bao-Liang Lu from Shanghai Jiao Tong University. 

The SEED database contains emotion-related EEG signals that are evoked by 15 film clips (with positive, neutral, and negative emotions) from 15 subjects with 3 sessions each. The signals are recorded by a 62-channel ESI neuroscan system.

The SEED-IV is an evolution of SEED, which contains 3 sessions, each has 15 subjects and 24 film clips. Comparing to the SEED with EEG signals only, this database also includes eye movement features recorded by SMI eye-tracking glasses.

\subsection{Pre-processing}\label{pre-processing}
After collecting EEG raw data, pre-processing on signals and feature extractions will be adopted. For both SEED and SEED-IV, to increase the SNR, the raw EEG signals are first down-sampled to a 200 Hz sampling rate, then been processed with a band-pass filter between 1 Hz to 75 Hz. After that, features are then being extracted.

\begin{equation}\label{DE}
\mathrm{DE}=-\int_{X} f(x) \log [f(x)] \mathrm{d} x
\end{equation}

Recent works extract features from EEG data on the time domain, frequency domain, and time-frequency domain. Among them, Differential Entropy (DE) as in \eqref{DE}, has the ability to distinguish patterns from different bands \cite{soleymani2015analysis}, thus we choose to take DE features as the input data of our model. For SEED and SEED-IV, extracted DE features at five frequency bands of delta (1-4 Hz), theta (4-8 Hz), alpha (8-14 Hz) and gamma (31-50 Hz) are provided.

One data from one subject in one session for both databases is in the form of channel (62) $\times$ trial (15 for SEED, 24 for SEED-IV) $\times$ band (5), we then merge the channel with the band, and the form becomes trial $\times$ 310 (62 $\times$ 5). For SEED, 15 trials contain 3394 samples in total for each session. For SEED-IV, 24 trials contain 851/832/822 samples for three sessions, respectively. In the end, all data are formed into 3394 $\times$ 310 (SEED), or 851/832/822 $\times$ 310 (SEED-IV) with corresponding generated label vectors in the form of 3394 $\times$ 1, or 851/832/822 $\times$ 1.

\section{Method}\label{method}
For simplicity of demonstration, we list the symbols and their definition in Table \ref{tab:notation_table} that will be used in the following sections.

\begin{table}[ht]
    \centering
    \caption{Notation Table}
    \label{tab:notation_table}
    \begin{tabular}{c l}
        \hline
        \textbf{symbol} & \textbf{definition} \\
        \hline
        $X$ & Instance set (matrix) \\
        $Y$ & Label set (matrix) \\
        $S$ & Source domain \\
        $T$ & Target domain \\ 
        $N$ & number of source domains \\
        $Q$ & Common feature \\
        $R$ & Domain-specific feature \\
        $\hat{Y}$ & Predicted label (matrix) \\
        $\phi, \Phi$  & Mapping function \\
        $\mathcal{H}$ & Reproducing kernel Hilbert space \\
        $CFE$ & Common feature extractor \\
        $DSFE$ & Domain-specific feature extractor \\
        $DSC$ & Domain-specific classifier \\
        $x$ & Feature vector \\
        $y$ & Label vector \\
        $q$ & Feature vector after CFE \\
        $r$ & Feature vector after DSFE \\
        $\hat{y}$ & Predicted label vector \\
        \hline
    \end{tabular}
\end{table}

\begin{figure*}
\setlength{\belowcaptionskip}{-0.cm}
    \centering
    \includegraphics[width=1\linewidth]{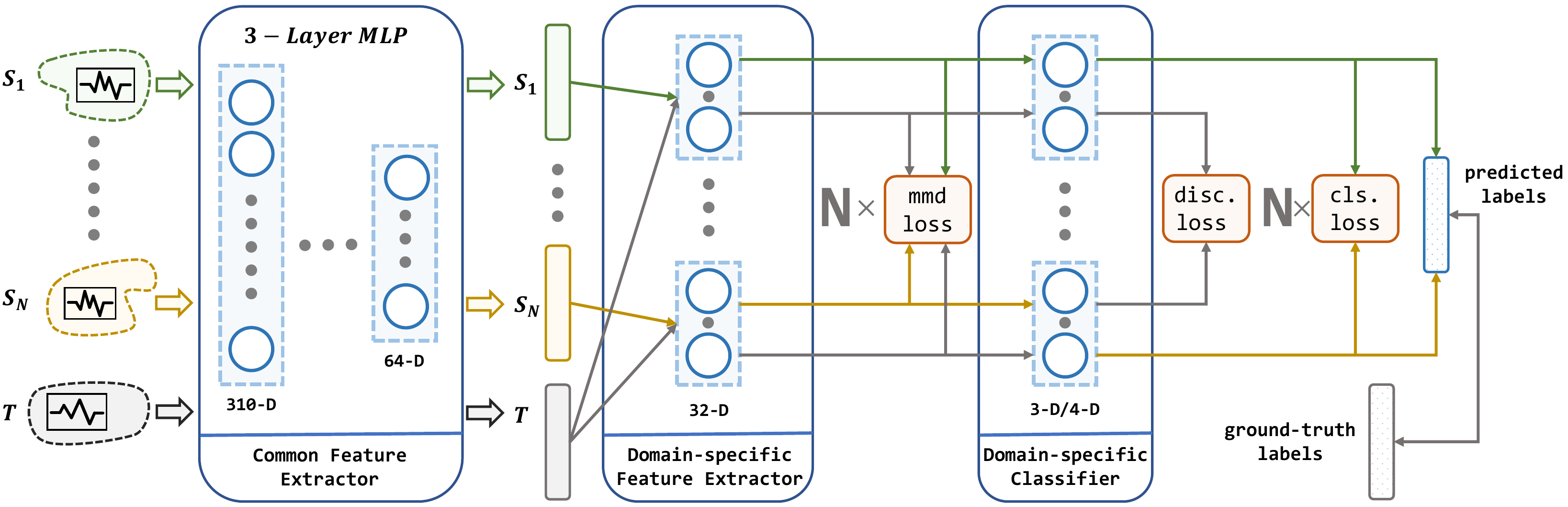}
    \caption{The {\bf architecture} of our proposed method. Our network consists of a common feature extractor, domain-specific feature extractor, and domain-specific classifier. For each source domain, a branch of DSFE and DSC is conducted for pair-wise domain adaptation. The model receives multiple source domains and leverages their knowledge to transfer to the target domain.}
    \label{fig:MS-MDA}
\end{figure*}

Given a set of pre-existing EEG data and a newly collected EEG data, our goal is to learn a model $\phi$ that is trained on these multiple independent source domain data using DA, and thus has a better prediction on the newly collected data than simply combining the existed data into one source domain. The architecture of the proposed method is illustrated in Fig. \ref{fig:MS-MDA}.

As shown in the figure, the input to the MS-MDA are $N$ independent source domain data $\{(\textbf{X}_{i}^{S}, \textbf{Y}_{i}^{S})\}_{i=1}^{N}$ and a target domain data $\{\textbf{X}^{T}\}$, and then these data are fed into a common feature extractor module to get the domain-invariance features $\{\textbf{Q}_{i}^{S}\}_{i=1}^{N}$ and $\{\textbf{Q}^{T}\}$. Then for each domain-specific feature extractor, extracted common features $\{\textbf{Q}_{i}^{S}\}_{i=1}^{N}$ will be fed into one branch with $\{\textbf{Q}^{T}\}$ and get their domain-specific features: $\{\textbf{R}_{i}^{S}\}_{i=1}^{N}$ and $\{\textbf{R}_{i}^{T}\}_{i=1}^{N}$, and on top of that, the MMD value is calculated, which is a measure of the distance of the current source and the target domain. Next, the target domain features $\{\textbf{R}_{i}^{T}\}_{i=1}^{N}$ and all the source domain features $\{\textbf{R}_{i}^{S}\}_{i=1}^{N}$ extracted from the last step will get to the domain-specific classifiers to get the corresponding classification predictions: $\{\hat{\textbf{Y}}^{T}_{i}\}_{i=1}^{N}$ and $\{\hat{\textbf{Y}}_{i}^{S}\}_{i=1}^{N}$, then the results of the source domain are taken to calculate the classification loss. Since the target domain will be fed into all the source domain classifiers, multiple target domain predictions are generated. These predictions are taken to calculate the discrepancy loss. In the end, the average of these target-domain predictions is taken as the output of the model. Details of these modules are given below.

{\bf Common Feature Extractor} in the MS-MDA is used to map the source and target domain data from the original feature spaces to a common sharing latent space, and then common representations of all domains are extracted. This module can help to extract some low-level domain-invariant features.

{\bf Domain-specific Feature Extractor} follows the Common Feature Extractor (CFE). After obtaining the features of all domains, we set up $N$ single fully connected layers to correspond to $N$ source domains. For each pair of source and target domain, we map the data to a unique latent space via the corresponding Domain-specific Feature Extractor (DSFE), respectively, and then obtain the domain-specific features in each branch. To apply DA and bring the two domains close in the latent space, we choose the MMD to estimate the distance between these two domains. MMD is widely used in the DA and can be formulated in \eqref{MMD}. In the process of training, MMD loss is decreased to narrow the source domain and the target domain in the feature space, which helps make better predictions for the target domain. This module aims to learn multiple domain-specific features. 

\begin{equation}\label{MMD}
\operatorname{MMD}_{\left(X^{S}, X^{T}\right)}=\left\|\frac{1}{N^{S}} \sum_{i=1}^{N^{S}} \Phi\left(\mathrm{x}_{i}^{S}\right)-\frac{1}{N^{T}} \sum_{j=1}^{N^{T}} \Phi\left(\mathrm{x}_{j}^{T}\right)\right\|_{\mathcal{H}}^{2}
\end{equation}

{\bf Domain-specific Classifier} uses the features extracted from the DSFE to predict the result. In Domain-specific Classifier (DSC), there are $N$ single softmax classifiers that correspond to each source domain. For each classifier training, we choose cross-entropy to estimate the classification loss, as shown in \eqref{clsloss}. Besides, since there are $N$ classifiers in this module, and these $N$ classifiers are trained on $N$ source domains, if their predictions are simply averaged as the final result, the variance will be high, especially when the target domain samples are at the decision boundary, which will have a significant negative impact on the results. To reduce this variance, a metric called discrepancy loss is introduced to make the predictions of the $N$ classifiers converge, which is shown in \eqref{discloss}. The average of the predictions of the $N$ classifiers is taken as the final result.

\begin{equation}\label{clsloss}
    \mathcal{L}_{c l s}=\sum_{i=1}^{N} \mathbf{E}_{x \sim X_{S}} J\left(\hat{\mathbf{Y}}_{i}^{S}, \mathbf{Y}_{i}^{S}\right)
\end{equation}

\begin{equation}\label{discloss}
    \mathcal{L}_{disc} = \sum_{i \neq j}^{N} \mathbf{E}_{x \sim X_{t}} \left|\hat{\textbf{Y}}_{i}^{T}-\hat{\textbf{Y}}_{j}^{T}\right|
\end{equation}

In summary, MS-MDA accepts $N$ source domain EEG data and one target domain EEG data, and then includes a common feature extractor to get $N$ source domain features and one target domain feature. Next, $N$ domain-specific feature extractors are used to pairwise compute the MMD loss of one individual source with the target domain and extract their domain-specific features. Finally, a domain-specific classifier is used to do the classification task, which also calculates the classification loss of the $N$ classifiers using the features, with the discrepancy loss of the $N$ classifiers for the features of the target domain data after the previous $N$ feature extractors.

\begin{equation}\label{totalloss}
    \mathcal{L} = \mathcal{L}_{cls} + \alpha \mathcal{L}_{mmd} + \beta \mathcal{L}_{disc}
\end{equation}

The training is based on the \eqref{totalloss} and following the algorithm as shown in Algorithm. \ref{alg:MS-MDA}. For the three losses, minimizing MMD loss can get domain-invariant features for each pair of the source and target domains; minimizing classification loss will bring more accurate classifiers for predicting the source domain data; minimizing discrepancy loss will get more convergent multiple classifiers.

\begin{algorithm}[ht]
\caption{Overview of MS-MDA} 
\label{alg:MS-MDA} 
\begin{algorithmic}[1]
\REQUIRE ~~ \\ 
Iteration $\mathcal{T}$, source domain data $\{(\textbf{X}_{i}^{S}, \textbf{Y}_{i}^{S})\}_{i=1}^{N}$ and target domain data $\{\textbf{X}^{T}\}$
\FOR{t = 1,..., $\mathcal{T}$}
    \STATE Take $m$ samples $\{x_{j}^{Si}, y_{j}^{Si}\}_{j=1}^{m}$from source domains and $\{x_{j}^{T}\}_{j=1}^{m}$ from target domain.
    \STATE $\{q_{j}^{Si}\}_{j=1}^{m}, \{q^{T}\} \leftarrow CFE(\{x_{j}^{Si}, y_{j}^{Si}\}_{j=1}^{m}, \{x_{j}^{T}\}_{j=1}^{m})$
    \STATE $\{r_{j}^{Si}\}_{j=1}^{m}, \{r_{j}^{T}\}_{j=1}^{m} \leftarrow DSFE(\{q_{j}^{Si}\}_{j=1}^{m}, \{q^{T}\})$
    \STATE $\mathcal{L}_{mmd} \leftarrow (\ref{MMD}) \leftarrow DSFE$
    \STATE $\{\hat{y}_{j}^{Si}\}_{j=1}^{m}, \{\hat{y}_{j}^{T}\}_{j=1}^{m}, \leftarrow DSC(\{r_{j}^{Si}\}_{j=1}^{m}, \{r_{j}^{T}\}_{j=1}^{m})$
    \STATE $\mathcal{L}_{cls}, \mathcal{L}_{disc} \leftarrow (\ref{clsloss})(\ref{discloss}) \leftarrow DSC$
    \STATE Update model by minimizing the total loss
\ENDFOR
\RETURN $\{\hat{Y}^{T}\}$;
\ENSURE ~~\\
Prediction of target domain data, $\{\hat{Y}^{T}\}$;
\end{algorithmic}
\end{algorithm}

\section{Experiments}\label{experiments}
We perform substantial experiments in the task of classification of emotions on two datasets SEED and SEED-IV, with the normalization study to the EEG data for domain adaptation. Besides, we also conduct some exploratory experiments in addition to the evaluation of our proposed methods and comparison methods.

\subsection{Implementation Details}
As mentioned in the Section. \ref{method}, there are many details in the three modules of MS-MDA. First, for the Common Feature Extractor (CFE), since we do not take raw data (\emph{i. e.} EEG signals) but the extracted DE features as vectors, complex deep models such as deep convolutional neural networks are not suitable for this module, thus we choose 3-layer MLP for simplicity which reduces feature dimensions from 310-dimension (62 $\times$ 5, channel $\times$ band) to 64-D. In CFE, every linear layer is followed by a LeakyReLU \cite{xu2015empirical} layer. We also evaluate the effort of the ReLU \cite{nair2010rectified} activation function, but due to the sensitivity of the EEG data, much information would be lost if using ReLU since the value less than zero would be dropped, so we choose LeakyReLU as a compromise. Next, for both domain-specific feature extractor (DSFE) and domain-specific classifier (DSC), there is a single linear which reduces 64-D to 32-D and 32-D to the corresponding number of categories (3 for SEED, 4 for SEED-IV), respectively. In DSFE, same as the settings in CFE, a LeakyReLU layer is followed after the linear layer, while in DSC, there is only one linear layer without any activation function. The network is trained using an Adam \cite{kingma2014adam} optimizer with an initial learning rate of 0.01, and train for 200 epoch. The batch size we choose is 256, which means we take 256 samples from each domain in every iteration (we also evaluate different settings of batch size and epoch in Section \ref{additions}). The whole model is trained under the \eqref{totalloss}, for domain adaptation loss, we choose MMD as the metric of the distance between two domains in the feature space (CORAL loss has a similar effect). As for the discrepancy loss, L1 regularization is being used, we also evaluate this loss in Section \ref{additions}. Besides, we dynamically adjust the $\alpha$ coefficients to achieve the effect of focusing on the classification results first, and then start aligning MMD and the convergence between the classifiers ($\alpha = \frac{2}{1+e^{-10*i/epoch}}-1$). As for the training data, we take the DE features and reform one sample to a 310-D vector as illustrated in the Section \ref{pre-processing}. Before feeding into the model, we normalize all the data in electrode-wise, refer to Section \ref{normalization} for details.

\subsection{Results}

\begin{table}[!t]
\centering
\caption{Comparison results on SEED and SEED-IV}
\label{tab:results}
\begin{tabular}{llll}
\hline
Dataset                  & Method                   & Cross-session                         & Cross-subject                         \\ \hline
\multirow{4}{*}{SEED}    & DDC                      & 81.53 \scriptsize $\pm$ 6.83          & 68.99 \scriptsize $\pm$ 3.23 \\
                         & DAN                      & 79.93 \scriptsize $\pm$ 7.06          & 65.84 \scriptsize $\pm$ 2.25          \\
                         & DAN \cite{li2018cross}               & -  & 83.81 \scriptsize $\pm$ 8.56
                         \\
                         & DCORAL                   & 76.86 \scriptsize $\pm$ 7.61          & 66.29 \scriptsize $\pm$ 4.53          \\
                         & DANN \cite{li2018cross}              & - & 79.19 \scriptsize $\pm$ 13.14 \\
                         & PPDA \cite{zhao2021plug}              & - & 86.70 \scriptsize $\pm$ 7.10  \\   
                         & \textbf{MS-MDA (Ours)} & \textbf{88.56 \scriptsize $\pm$ 7.80} & \textbf{89.63 \scriptsize $\pm$ 6.79}          \\ \hline
\multirow{4}{*}{SEED-IV} & DDC                      & 57.63 \scriptsize $\pm$ 11.28         & 37.41 \scriptsize $\pm$ 6.36          \\
                         & DAN                      & 55.14 \scriptsize $\pm$ 12.79         & 32.44 \scriptsize $\pm$ 9.02          \\
                         & DCORAL                   & 44.63 \scriptsize $\pm$ 11.38         & 37.43 \scriptsize $\pm$ 3.08          \\
 & \textbf{MS-MDA (Ours)} & \textbf{61.43 \scriptsize $\pm$ 15.71} & \textbf{59.34 \scriptsize $\pm$ 5.48} \\ 
\hline 
\end{tabular}
\end{table}

Experiment results of comparison methods and our proposed method on SEED and SEED-IV are listed in Table. \ref{tab:results}, all the hyper-parameters are the same, except for those results taken directly from the original papers. It should be noticed that since many previous works do not make their codes public available, we then customize the comparison methods (in the deep learning domain adaptation field) that are described in their papers with our settings, and also including some typical deep learning domain adaptation models for better comparison (DDC \cite{tzeng2014deep}, DCORAL \cite{sun2016deep}). The results indicate that our method largely outperforms the comparison methods in most transfer scenarios. For SEED dataset, our method has a minimum of ~7\% and ~3\% improvement in cross-session and cross-subject scenarios, respectively. While in SEED-IV dataset, our method has a minimum of ~7\% and ~18\% for two transfer scenarios. The results also show that our method outperforms comparison methods significantly in cross-subject, the reason for that may be that in the cross-subject scenario, the number of sources is 14, much bigger than the number of 2 in cross-session, and thus maximizes the effect of taking multiple sources as multiple individuals in domain adaptation rather than concatenating them.

\begin{table}[!t]
\centering
\caption{Ablation study of MS-MDA on SEED and SEED-IV}
\label{tab:ablation_study}
\begin{tabular}{llll}
\hline
Dataset                  & Method               & Cross-session & Cross-subject \\ \hline
\multirow{4}{*}{SEED}    & Ours full            & \textbf{88.56 \scriptsize $\pm$ 7.80} & \textbf{89.63 \scriptsize $\pm$ 6.79} \\ \cline{2-4} 
                         & w/o MMD loss         & 82.20 \scriptsize $\pm$ 14.33 & 67.65 \scriptsize $\pm$ 17.65 \\
                         & w/o disc. loss       & 86.27 \scriptsize $\pm$ 9.14 & 87.27 \scriptsize $\pm$ 5.70 \\
                         & w/o MMD + disc. loss & 83.19 \scriptsize $\pm$ 9.69 & 80.48 \scriptsize $\pm$ 5.76 \\ \hline
\multirow{4}{*}{SEED-IV} & Ours full            & 61.43 \scriptsize $\pm$ 15.71 & \textbf{59.34 \scriptsize $\pm$ 5.48} \\ \cline{2-4} 
                         & w/o MMD loss         & 56.51 \scriptsize $\pm$ 18.48 & 49.71 \scriptsize $\pm$ 5.82 \\
                         & w/o disc. loss       & 61.63 \scriptsize $\pm$ 17.62 & 55.37 \scriptsize $\pm$ 14.38 \\
                         & w/o MMD + disc. loss & \textbf{62.66 \scriptsize $\pm$ 16.07} & 55.81 \scriptsize $\pm$ 4.17 \\ \hline
\end{tabular}
\end{table}

\subsection{Ablation Study}\label{ablation}
To understand the effect of each module in the MS-MDA, we remove them one at a time and evaluate the performance of the ablated model, the results are shown in Table. \ref{tab:ablation_study}. The first row of SEED and SEED-IV shows the performance of the full model (the same as in Table \ref{tab:results}). The second row ablates the MMD loss in the training process, which makes the model focuses only on the classification loss and discrepancy loss. The significant drop compared to the full model indicates the important effect of domain adaptation. Notice that even the results without MMD loss are better than many comparison methods, showing the importance of taking multiple sources as multiple individuals during training. The third row of taking out the discrepancy loss shows that this loss will affect the performance but the impact is minimal, the reason is that we want this discrepancy loss to be the icing on the cake rather than having a dominant effect on the model. The fourth row only considers the classification loss, thus reduces losses \eqref{MMD} and \eqref{discloss}.

\subsection{Normalization}\label{normalization}

\begin{figure}[!t]
\centering
  \begin{minipage}[t]{0.45\linewidth}
    \centering
    \includegraphics[scale=0.5]{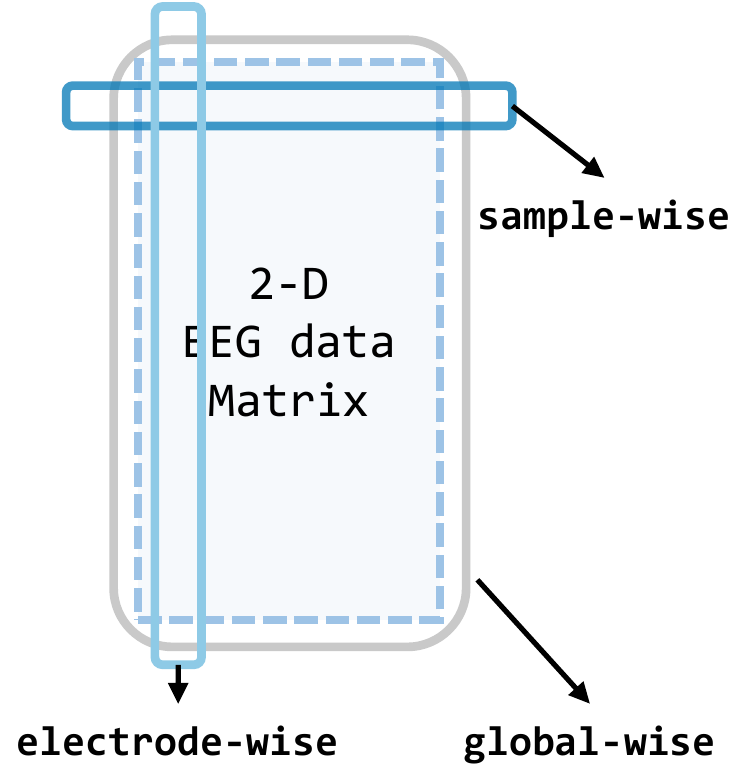}
    \caption{Three normalization methods. The dark blue box stands for the sample-wise normalization, while the light blue box stands for the electrode-wise normalization. The big gray box stands for the global-wise normalization.}
    \label{fig:normalization_type}
  \end{minipage}
  \begin{minipage}[t]{0.45\linewidth}
    \centering 
    \includegraphics[scale=0.5]{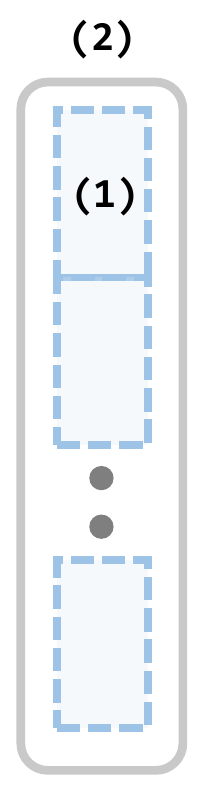}
    \caption{Small blue matrices are data from different subjects, and (1), (2) are two operations. The basic process is: multi-source data \textrightarrow (1) \textrightarrow (2). In order A, (1) in the figure stands for the normalization, and (2) stands for the concatenate. In order B: (1) stands for concatenating while (2) is for normalization.}
    \label{fig:normalization_order}
  \end{minipage}
\end{figure}

\begin{table*}[!t]
\centering
\caption{Normalization Study of MS-MDA and DAN. DAN$^{1}$ stands for the order A while DAN$^{2}$ stands for the order B.}
\label{tab:normalization_study}
\begin{tabular}{llllll}
\hline
\multirow{2}{*}{Model} & \multirow{2}{*}{Normalization type} & \multicolumn{2}{c}{SEED} & \multicolumn{2}{c}{SEED-IV} \\ \cline{3-6} 
                          &                   & Cross-session & Cross-subject & Cross-session & Cross-subject \\ \hline
\multirow{4}{*}{DAN$^{1}$}  & w/o normalization & 33.96 \scriptsize $\pm$ 0.23 & 33.91 \scriptsize $\pm$ 0.09 & 27.23 \scriptsize $\pm$ 4.78 & 27.15 \scriptsize $\pm$ 1.31 \\
                          & electrode-wise    & \textbf{79.93 \scriptsize $\pm$ 7.06} & \textbf{65.84 \scriptsize $\pm$ 2.25} & \textbf{55.14 \scriptsize $\pm$ 12.79} & \textbf{32.44 \scriptsize $\pm$ 9.02} \\
                          & sample-wise       & 52.51 \scriptsize $\pm$ 11.92 & 51.77 \scriptsize $\pm$ 12.61 & 27.34 \scriptsize $\pm$ 2.45 & 32.03 \scriptsize $\pm$ 4.24 \\
                          & global-wise       & 54.02 \scriptsize $\pm$ 9.29 & 49.12 \scriptsize $\pm$ 12.06 & 31.72 \scriptsize $\pm$ 6.46 & 29.31 \scriptsize $\pm$ 2.40 \\ \hline
\multirow{4}{*}{DAN$^{2}$}  & w/o normalization & 33.96 \scriptsize $\pm$ 0.23 & 33.91 \scriptsize $\pm$ 0.09 & 27.23 \scriptsize $\pm$ 4.78 & 27.15 \scriptsize $\pm$ 1.31 \\
                          & electrode-wise    & \textbf{79.78 \scriptsize $\pm$ 6.97} & \textbf{62.57 \scriptsize $\pm$ 5.31} & \textbf{52.18 \scriptsize $\pm$ 10.53} & \textbf{34.26 \scriptsize $\pm$ 7.98} \\
                          & sample-wise       & 52.51 \scriptsize $\pm$ 11.92 & 51.77 \scriptsize $\pm$ 12.61 & 27.34 \scriptsize $\pm$ 2.45 & 32.03 \scriptsize $\pm$ 4.24 \\
                          & global-wise       & 53.07 \scriptsize $\pm$ 10.50 & 50.22 \scriptsize $\pm$ 3.66 & 31.01 \scriptsize $\pm$ 7.56 & 31.77 \scriptsize $\pm$ 2.08 \\ \hline
\multirow{4}{*}{MS-MDA} & w/o normalization & 80.62 \scriptsize $\pm$ 12.22 & 60.92 \scriptsize $\pm$ 3.58 & 30.11 \scriptsize $\pm$ 6.47 & 29.64 \scriptsize $\pm$ 7.26 \\
                          & electrode-wise    & \textbf{86.94 \scriptsize $\pm$ 8.68} & \textbf{86.93 \scriptsize $\pm$ 8.24} & \textbf{64.07 \scriptsize $\pm$ 14.36} & \textbf{55.21 \scriptsize $\pm$ 6.30} \\
                          & sample-wise       & 81.84 \scriptsize $\pm$ 13.72 & 74.09 \scriptsize $\pm$ 5.79 & 34.71 \scriptsize $\pm$ 10.94 & 30.25 \scriptsize $\pm$ 5.20 \\
                          & global-wise       & 81.80 \scriptsize $\pm$ 12.75 & 78.89 \scriptsize $\pm$ 10.38 & 33.87 \scriptsize $\pm$ 10.99 & 31.88 \scriptsize $\pm$ 7.73 \\ \hline
\end{tabular}
\end{table*}

\begin{table*}[!t]
\centering
\caption{Performance of MS-MDA on SEED and SEED-IV with different settings of $\beta$. Training percentage stands for when to add this loss into the training, 1 means whole training process while 0.2 stands for the last 20\% of the training process. Weight represents the ratio compared to $\alpha$.}
\label{tab:beta_study}
\begin{tabular}{ccllll}
\hline
\multicolumn{1}{c}{\multirow{2}{*}{training percentage}} &
  \multicolumn{1}{c}{\multirow{2}{*}{weight}} &
  \multicolumn{2}{c}{SEED} &
  \multicolumn{2}{c}{SEED-IV} \\ \cline{3-6} 
\multicolumn{1}{c}{} &
  \multicolumn{1}{c}{} &
  \multicolumn{1}{l}{Cross-session} &
  \multicolumn{1}{l}{Cross-subject} &
  \multicolumn{1}{l}{Cross-session} &
  \multicolumn{1}{l}{Cross-subject} \\ \hline
\multicolumn{2}{c}{w/o disc. loss} & 86.27 \scriptsize $\pm$ 9.14 & 87.27 \scriptsize $\pm$ 5.70 & 61.63 \scriptsize $\pm$ 17.62 & 55.37 \scriptsize $\pm$ 14.38 \\ \hline
0.2   & 1     & 86.94 \scriptsize $\pm$ 8.68 & 86.93 \scriptsize $\pm$ 8.24 & 64.07 \scriptsize $\pm$ 14.36 & 55.21 \scriptsize $\pm$ 6.30 \\
0.2   & 0.1   & 86.99 \scriptsize $\pm$ 9.23 & 87.37 \scriptsize $\pm$ 7.64 & \textbf{64.75 \scriptsize $\pm$ 13.36} & 53.15 \scriptsize $\pm$ 11.88 \\
0.2   & 0.01  & 86.87 \scriptsize $\pm$ 9.30 & 87.09 \scriptsize $\pm$ 8.02 & 64.04 \scriptsize $\pm$ 13.72 & 50.54 \scriptsize $\pm$ 15.59 \\
0.2   & 0.001 & 86.91 \scriptsize $\pm$ 9.35 & 86.93 \scriptsize $\pm$ 8.24 & 64.14 \scriptsize $\pm$ 13.88 & 53.25 \scriptsize $\pm$ 9.55 \\
1     & 1     & 85.58 \scriptsize $\pm$ 8.19 & 63.42 \scriptsize $\pm$ 2.15 & 61.88 \scriptsize $\pm$ 16.71 & 57.34 \scriptsize $\pm$ 9.07 \\
1     & 0.1   & 85.80 \scriptsize $\pm$ 10.05 & 81.13 \scriptsize $\pm$ 11.19 & 62.42 \scriptsize $\pm$ 15.99 & 56.34 \scriptsize $\pm$ 10.18 \\
1     & 0.01  & \textbf{88.56 \scriptsize $\pm$ 7.80} & \textbf{89.63 \scriptsize $\pm$ 6.79} & 61.43 \scriptsize $\pm$ 15.71 & \textbf{59.34 \scriptsize $\pm$ 5.48} \\
1     & 0.001 & 86.36 \scriptsize $\pm$ 8.68 & 84.84 \scriptsize $\pm$ 3.49 & 64.41 \scriptsize $\pm$ 17.58 & 48.01 \scriptsize $\pm$ 8.66 \\ \hline
\end{tabular}
\end{table*}

During the experiments, we also find that different normalization to data can significantly impact the outcomes, and also the order of whether first concatenating multiple sources or first normalize each session individually. Thus we design diagrams and conduct extensive experiments to investigate the effects of different normalization strategies on the input data, \emph{i. e.,} extracted feature vectors from two datasets. Since we have reformed the origin 4-D matrices (session $\times$ channel $\times$ trial $\times$ band) into 3-D matrices (session $\times$ trial $\times$ (channel*band)), for each session, there is a 2-D matrix of trial $\times$ 310. Following the common machine learning normalization approaches and the prior knowledge and intuition of EEG data (\emph{i. e.,} the data acquired by the same electrode are more consistent with the same distribution), the normalization methods to these 2-D matrices can be categorized into three, as shown in Fig. \ref{fig:normalization_type}. Besides, since we also take the multi-source situation into consideration, the order of normalization may also influence the performance, as shown in Fig. \ref{fig:normalization_order}.

We evaluate three normalization methods and two normalization orders on SEED and SEED-IV with our proposed method MS-MDA and representative domain adaptation model DAN \cite{long2015learning}. The results are listed in Table. \ref{tab:normalization_study}. In all three sets, the normalization of electrode-wise outperforms the other three normalization types significantly. Comparing DAN$^{1}$ with DAN$^{2}$, the results indicate that the first normalization order of normalizing the data first and then concatenating them is better. In the third set of MS-MDA, we find that all the results of four normalization types are better than those in the first and second sets, and the improvement is significant. Row w/o normalization in MS-MDA, for example, has a top of ~47\% improvement, which also indicates the generalization of our proposed method in different normalization types, and the positive effects of taking multiple sources as individual branches for DA.

\subsection{Additions}\label{additions}
\subsubsection{Coefficient Study}

After multiple sets of experiments, we find that easy to control the MMD loss and it plays an influential role in the training as shown in Table \ref{tab:ablation_study}. However, for the disc. loss, it remains many problems. Adding this loss to the model too early will affect the overall effect, and too late will lose the impact of learning convergence. Too large a weight would cause the training to focus on convergence, thus the few correct ones might follow the many incorrect ones; too small may not have enough influence on the model. Also, for better use and simplicity mentioned earlier, we do not make many tests on the $\beta$, but simply compared the effects on only a few sets of $\beta$, and the results are shown in Table \ref{tab:beta_study}. From which we can see that compared to row one (w/o disc. loss), introducing discrepancy loss increases the performance in most cases, especially when training for the whole process in cross-subject for SEED-IV. We then choose the weight of 0.01 and training discrepancy loss for the whole process according to the results.

\subsubsection{Hyper-parameters and Data Visualization}
To better investigating our proposed method, we evaluate it with different hyper-parameters, besides, we also take the representative method DAN as the comparison. The results are shown in Fig. \ref{fig:bs_epoch_1} and Fig. \ref{fig:bs_epoch_2}. From them we can see that, with the increase of batch size, both models show a drop in performance, especially when the batch size is 512, which has a significant decrease compared to 256 on SEED-IV. Besides, with the training epoch increases, neither model has a substantial improvement, especially MS-MDA, but our method achieves moderate accuracy and converges faster. Comparing cross-subject experiments on two datasets, it can be significantly seen that MS-MDA has a clear advantage over DAN, which indirectly shows that our approach has a more significant performance improvement for multiple source domain adaptation in EEG-based emotion recognition.

For a better understanding of the effect of our proposed method, we randomly pick 100 EEG samples from each subject (domain) in the scenario of cross-subject to visualize with t-SNE~\cite{van2008visualizing}, as displayed in Fig.~\ref{tsne}. We only plot the cross-subject since this transfer scenario has more sources that will maximize visualization. In the Fig.~\ref{tsne}, each color stands for a source domain, and the target domain are in black. To better plotting, we transparent the target sample to avoid overlap. It should be noticed that in the lower left figure, we pick 1400 samples since we concatenate all sources into one.

\begin{figure}[t]
    \centering
    \subfloat[MS-MDA with different batch size\label{fig:bs_msmda}]{\includegraphics[width=0.24\textwidth]{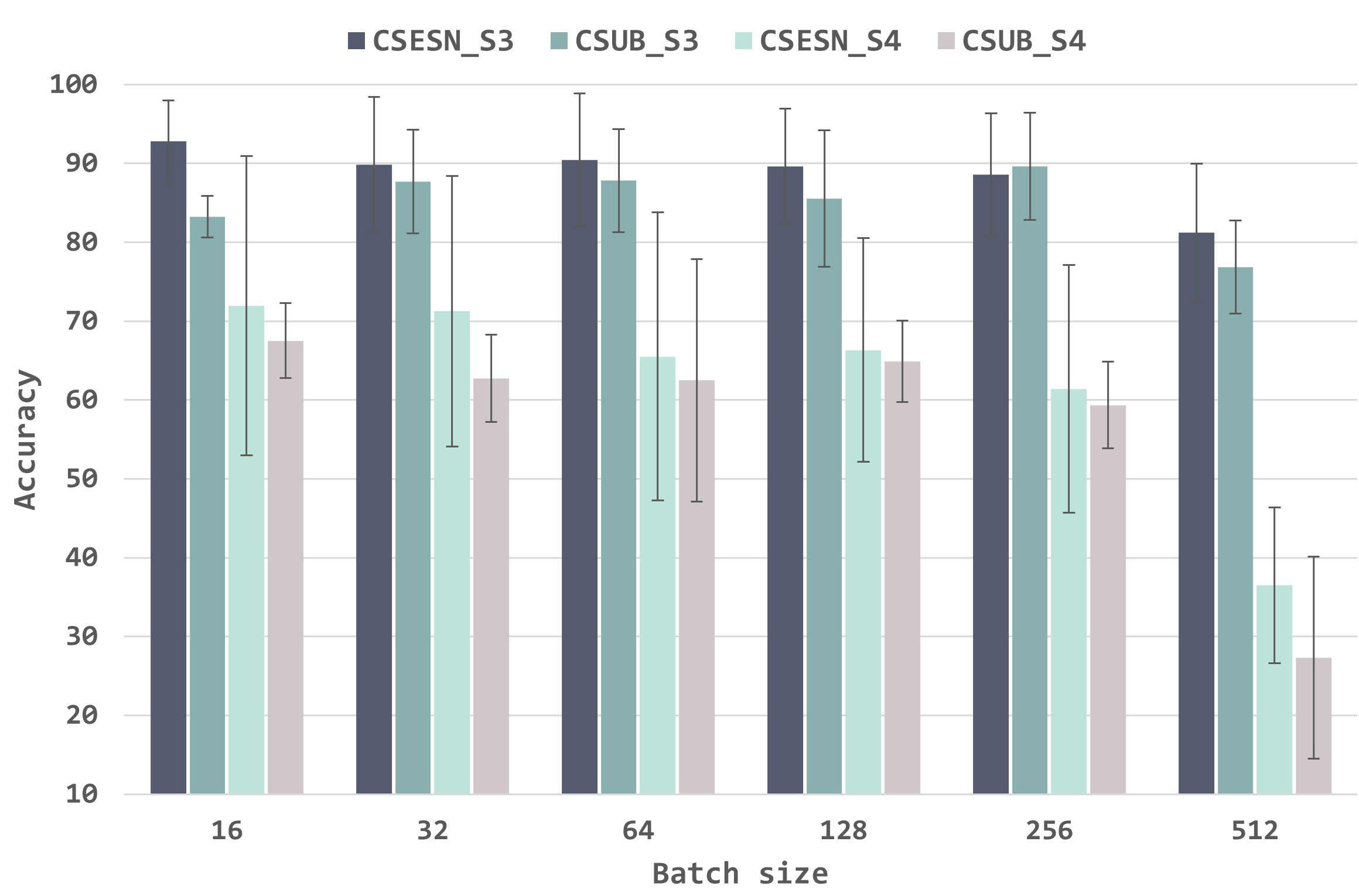}}
    \subfloat[DAN with different batch size\label{fig:bs_dan}]{\includegraphics[width=0.24\textwidth]{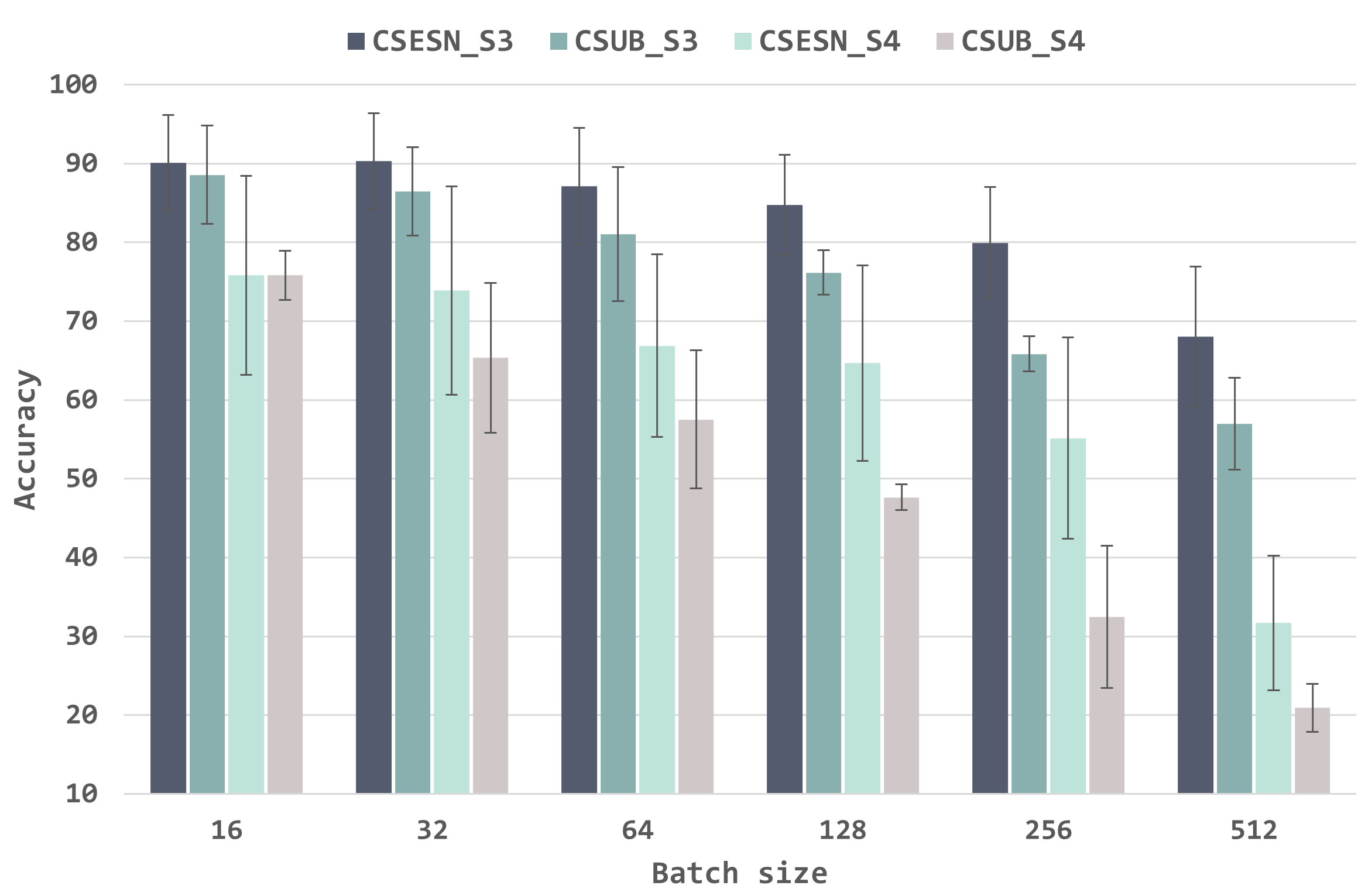}}
    \newline
    \subfloat[MS-MDA with different epochs\label{fig:epoch_msmda}]{\includegraphics[width=0.24\textwidth]{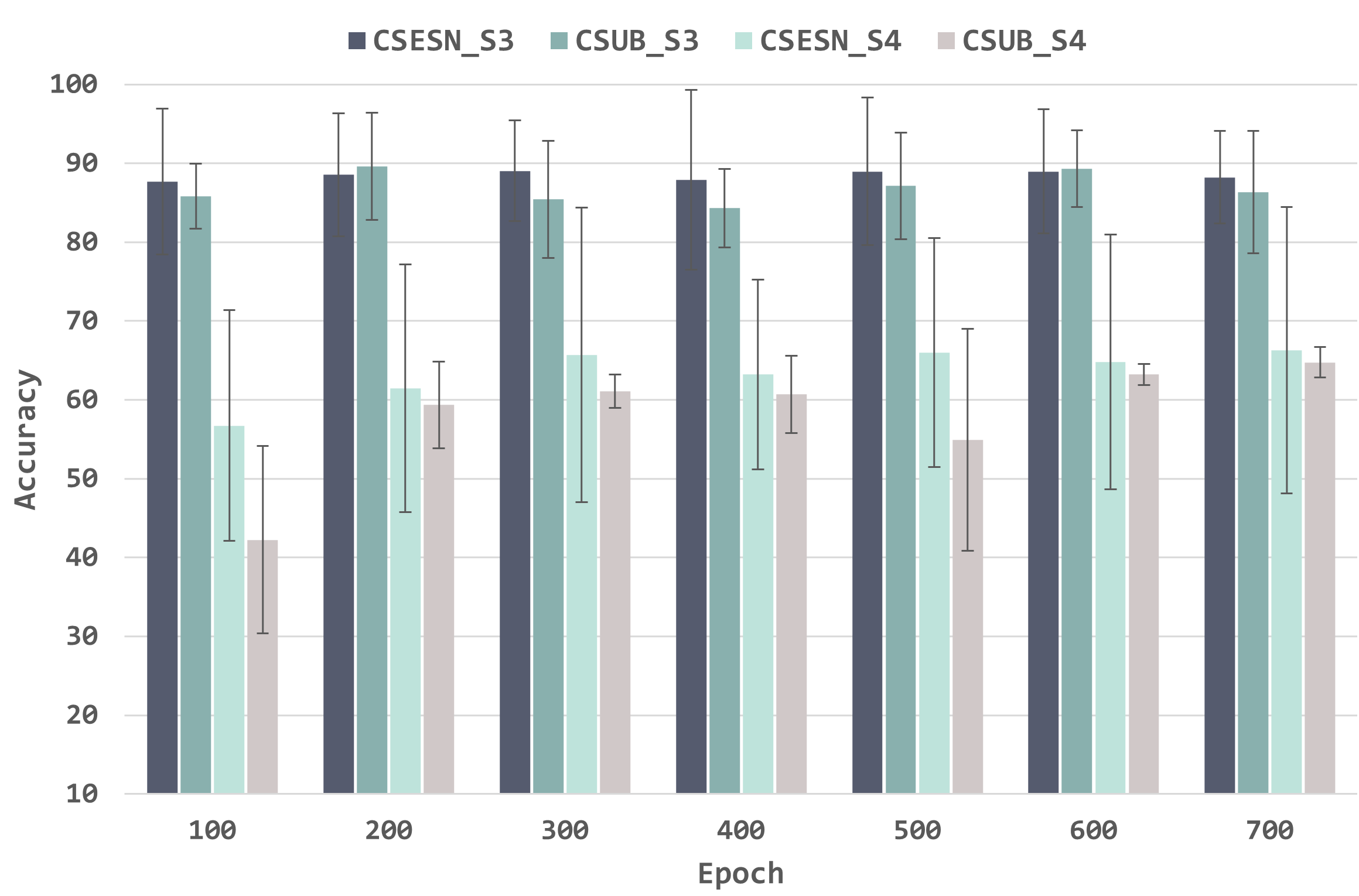}}
    \subfloat[DAN with different epochs\label{fig:epoch_dan}]{\includegraphics[width=0.24\textwidth]{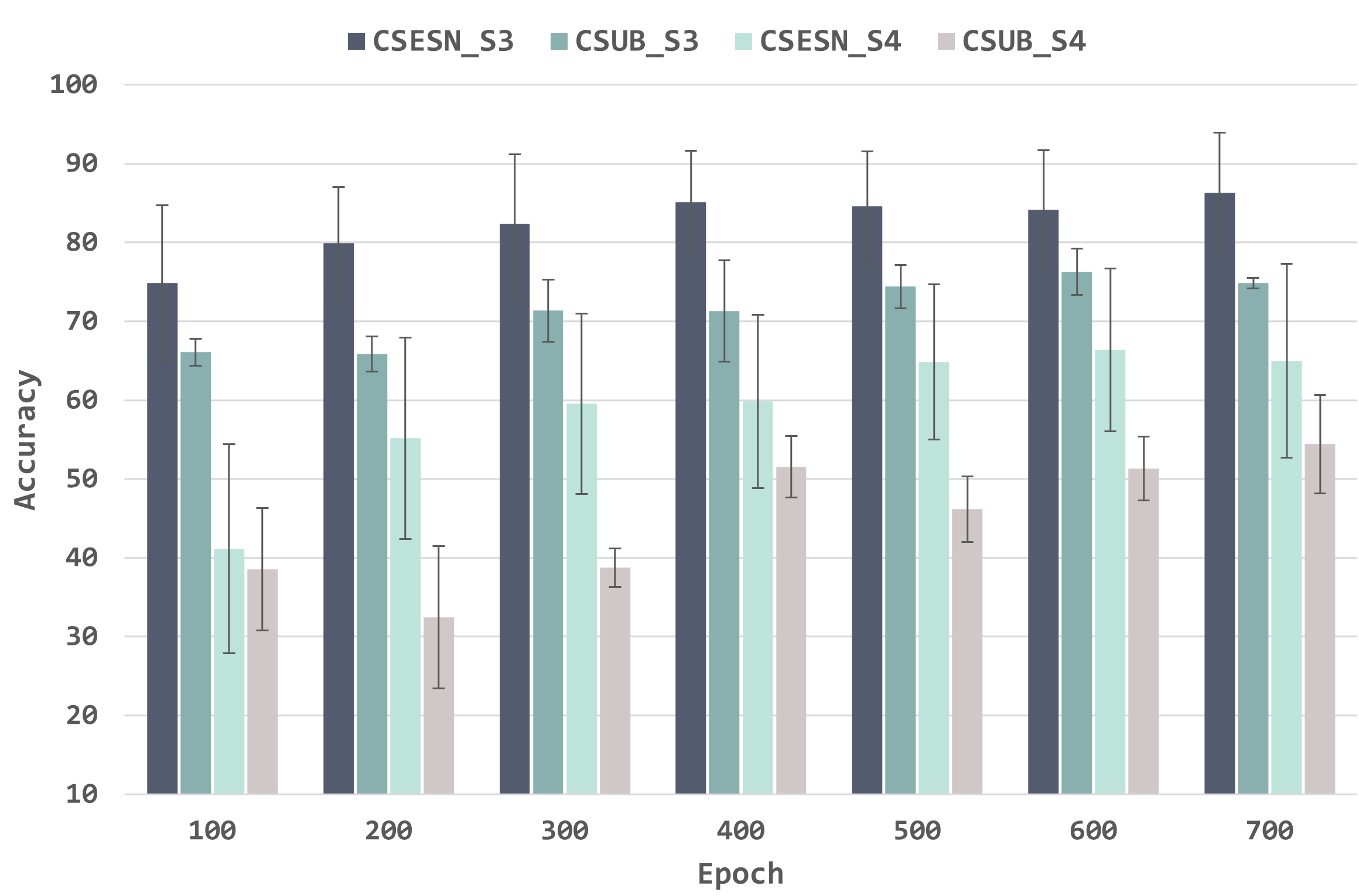}}
    \caption{Evaluation of MS-MDA and DAN with different batch size and epochs. Each bar stands for one cross scenario for one dataset.}
    \label{fig:bs_epoch_1}
\end{figure}

\begin{figure}[t]
    \centering
    \subfloat[Batch size\label{fig:bs}]{\includegraphics[width=0.24\textwidth]{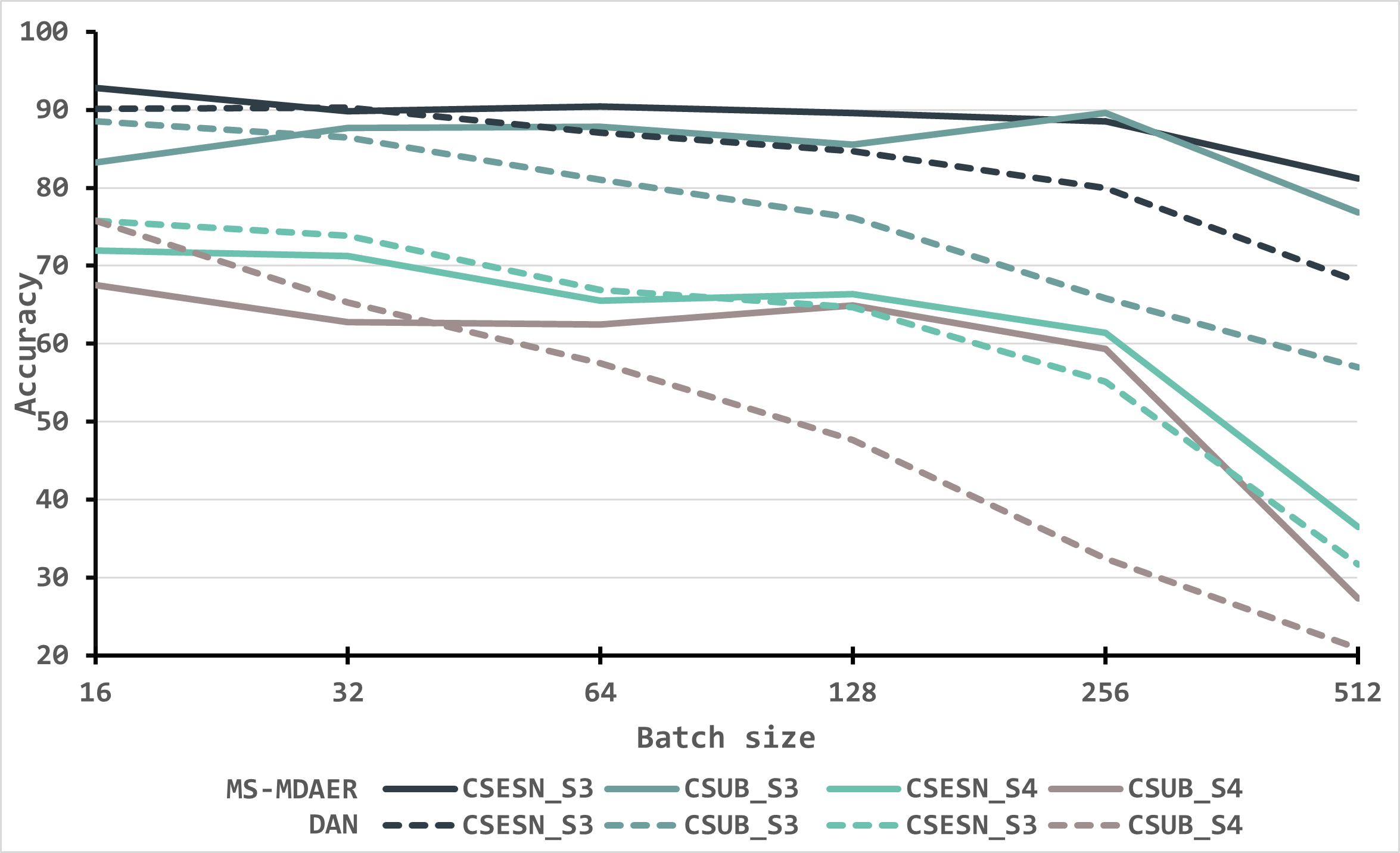}}
    \subfloat[Epoch\label{fig:epoch}]{\includegraphics[width=0.24\textwidth]{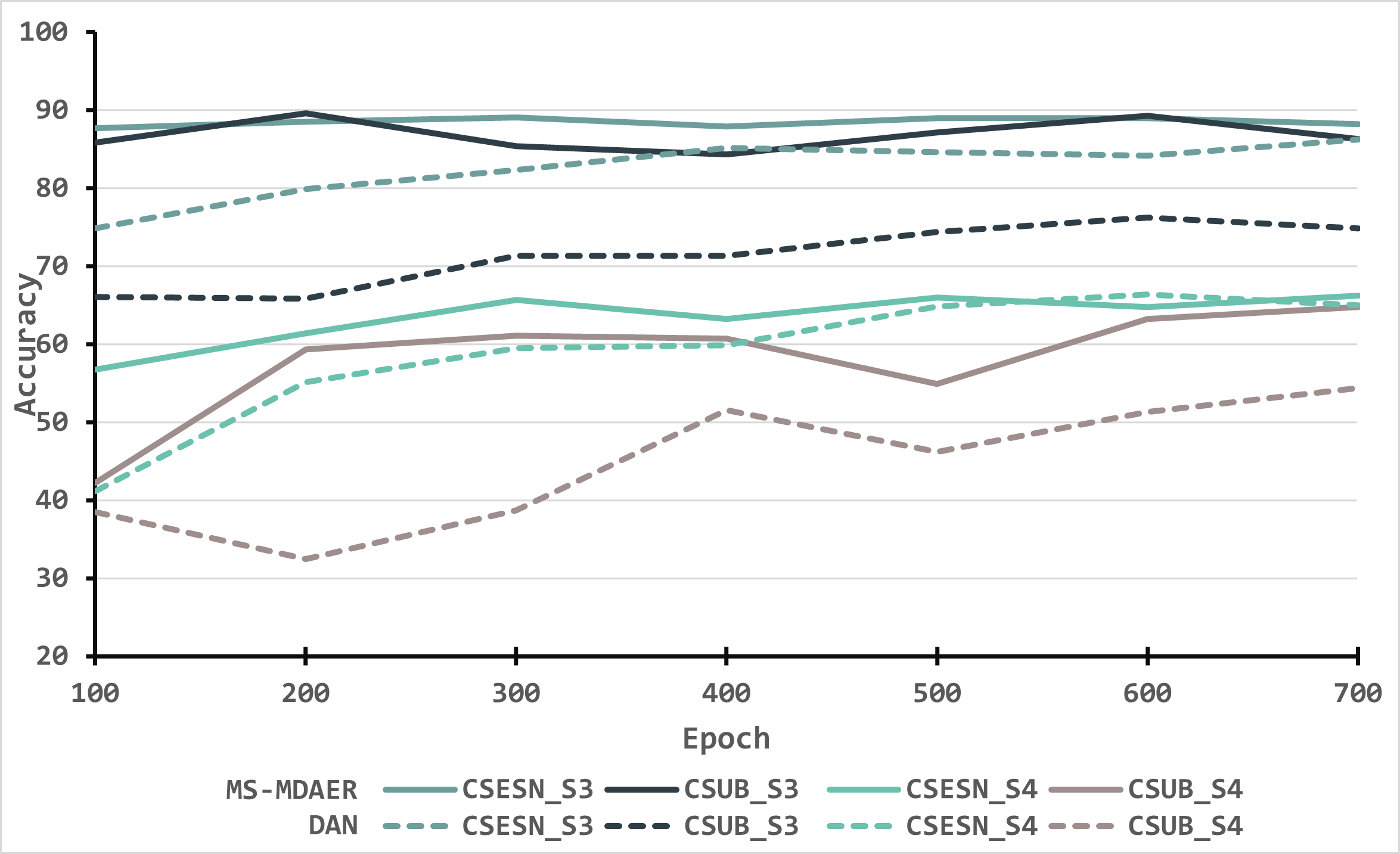}}
    \caption{Evaluation of MS-MDA and DAN with different settings of batch size and epochs. Each line stands for one cross scenario for one dataset.}
    \label{fig:bs_epoch_2}
\end{figure}
 
\begin{figure}[!t]
\centerline{\includegraphics[width=\columnwidth]{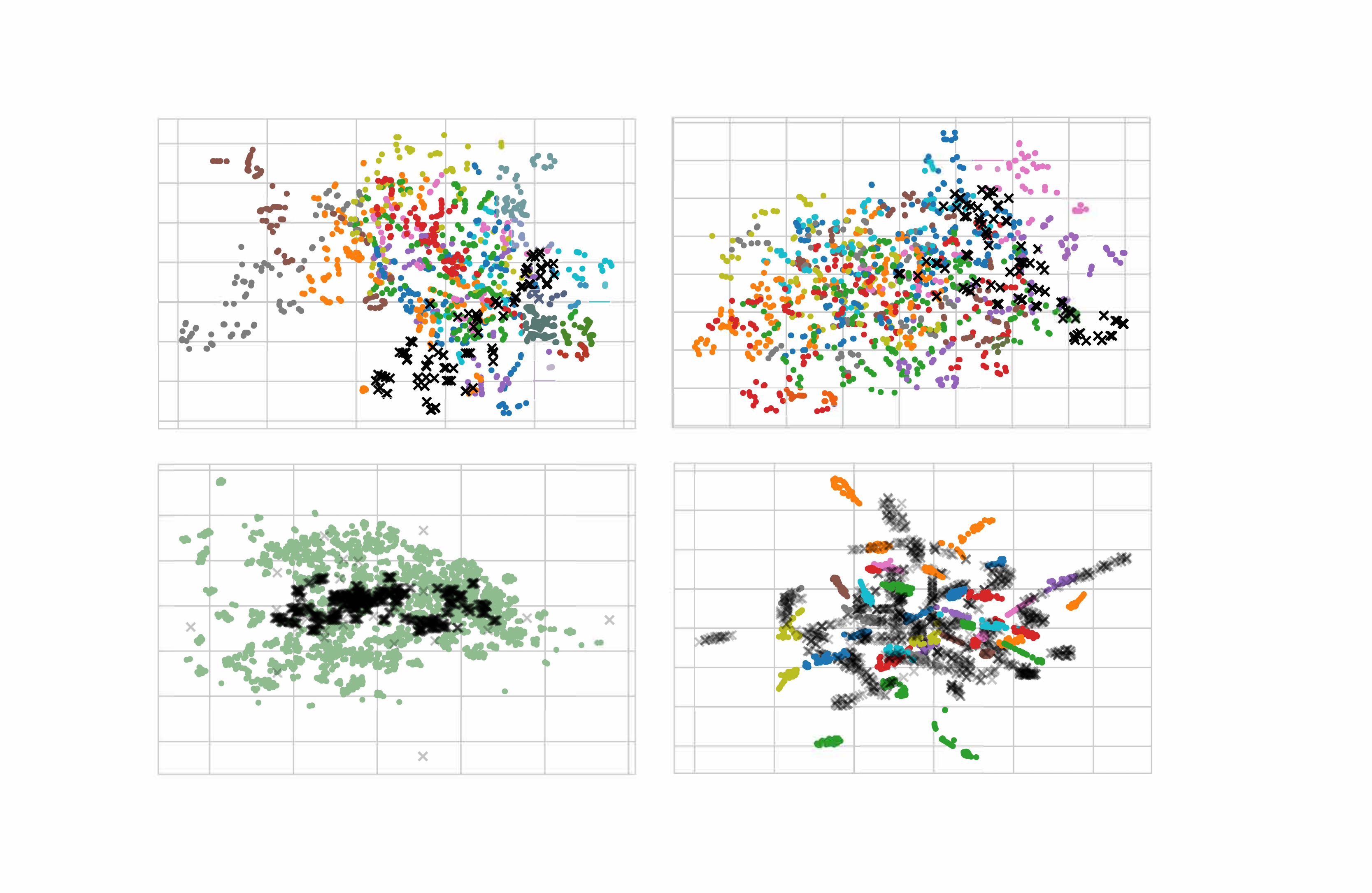}}
\caption{Visualization with t-SNE for raw data (\textbf{upper left}), normalization data (\textbf{upper right}), data using DAN (\textbf{lower left}), and data using MS-MDA (\textbf{lower right}). The input data of the last fully-connected (DSC) layer are used for the computation of the t-SNE. Target data are in the shape of X with black, all other 14 source data are in 14 colors. Notice that since we have concatenated all the source domains, the lower left figure has only one color for the source domain. All four figures are best viewed in color.}
\label{tsne}
\end{figure}

\section{Discussion}\label{discussion}
As can be seen from Table \ref{tab:results}, comparing the results of selective methods and prior works, our proposed method has a significant improvement, especially for cross-subject DA in which the number of source domains is large. The ablation experiments from Table \ref{ablation} also show that our proposed method requires both MMD and discrepancy loss in most cases. Eliminating the MMD loss has a significant performance drop on both datasets, confirming the importance of DA, and eliminating disc. loss does not have as large an impact as MMD loss, but also verifies the help of multi-source convergence. Also, during the experiments, we find that the type of normalization of the data has a significant impact on the overall results, so we also design experiments and explore the normalization of EEG data in DA to help improve the performance of our model. As can be seen in Table \ref{normalization}, there is not much difference between the two normalization orders, and it is most appropriate to do data normalization on the electrode-wise, which has a crushing performance improvement compared to the other three methods; for our method, which does not concatenate data, electrode normalization is also the most effective. This conclusion is in line with our intuition that data collected from the same electrode are relatively more regular or conform to a certain distribution, while data collected from different electrodes are very different. In addition, during the experiments, we find that the disc. loss needs to be carefully adjusted, otherwise it is easy to cause harmful effects, which we guess is because this loss introduces a convergence effect on multiple classifiers in the model (in other words, smooth the inferences made from multiple classifiers), and if most of the classifiers are wrong, this convergence effect will cause the correct classifiers to error. Therefore, we also test and evaluate the impact of the disc. loss coefficients on the model at different settings, and from Table \ref{tab:beta_study}, we can see that the disc. loss achieves the best results if it is set to 0.01 times the MMD loss coefficient and is being used in the full model training.

After exploring the internal details of the model, we also evaluated the performance of the model under different hyper-parameters. For better comparison, we chose a representative DAN as the comparison method. From Fig. 6 and Fig. 7, we can see that both models have a significant decrease as the batch size is increased. The reason for this we assume is that small batch size tends to fall into local optimal overfitting. The performance of both models increases slightly with epoch. From Figs. 6 and 7, we can also clearly see that MS-MDA has a significant advantage over DAN in cross-subject DA where the number of multiple source domains is large, which also confirms the importance of constructing multiple branches for multiple source domains to adopt DA separately.

Although it is clear from the results that our proposed method has a significant performance improvement, we also found that the training time consumed increases linearly with the number of source domains, i.e., the larger the number of source domains and the larger the model, the longer the training takes, unlike concatenating all source data into one, where there is only additional time due to the increase in the amount of data. For this problem, our current idea is to discard some less relevant source domains selectively and not build DA branches for them, allowing the disc. loss to play a more prominent role because there is less negative information. In addition, the encoders in the current model are the simplest MLP, and many literature and works have verified the usability of LSTM for EEG data \cite{jiao2020driver, tao2020emotion, ma2019emotion}, and we will consider switching to use LSTM as the encoders in future works.

\section{Conclusion}\label{conclusion}
In this paper, we propose MS-MDA, an EEG-based emotion recognition domain adaptation method, which is applicable to multiple source domain situations. Through experimental evaluation, we find that this method has a better ability to adapt to multiple source domains, which is validated by comparison with the selective approaches and the SOTA models, especially for cross-subject experiments where our proposed method consists of up to 20\% improvement. In addition, we also explore the impact of different normalization methods for EEG data in domain adaptation, which we believe can serve as an inspiration for other EEG-based works while improving the effectiveness of the models. As for our future work, the current model for multiple source domains is to construct a DA branch for each of them without selection, which will increase the model size and training time exponentially, and also introduces information from the source domain that is not relevant to the target into the model. A more efficient approach may be to selectively build DA branches from a reservoir of source domains, allowing the model to be more efficient while only focusing on the source domain information that is relevant to the target domain.
% \newpage
\balance
\bibliographystyle{IEEEtran}
\bibliography{main}
\end{document}